\documentclass[sigconf]{acmart}
\AtBeginDocument{%
  }
\usepackage{algorithmic}

\setcopyright{acmlicensed}
\copyrightyear{2018}
\acmYear{2018}
\acmDOI{XXXXXXX.XXXXXXX}
\acmConference[Conference acronym 'XX]{Make sure to enter the correct
  conference title from your rights confirmation email}{June 03--05,
  2018}{Woodstock, NY}
\acmISBN{978-1-4503-XXXX-X/2018/06}

\usepackage[ruled,linesnumbered]{algorithm2e}
\usepackage{amsmath}

\usepackage{amssymb}

\begin{document}

\title{Audio-Driven Adversarial Defense for 3D Talking Face Generation with totally Visual Fidelity Preservation}

\author{Rui-Qing Sun}
\email{2325557558@qq.com}
\affiliation{%
  \institution{Beijing Institute of Technology}
  \city{Haidian Qu}
  \state{Beijing Shi}
  \country{China}
}

\author{Chen-Hao Cui}
\affiliation{%
  \institution{Beijing Institute of Technology}
  \city{Haidian Qu}
  \state{Beijing Shi}
  \country{China}
}

\author{Hui-Yang Zhao}
\affiliation{%
  \institution{Beijing Institute of Technology}
  \city{Haidian Qu}
  \state{Beijing Shi}
  \country{China}
}

\author{Tian Lan}
\affiliation{%
  \institution{Beijing Institute of Technology}
  \city{Haidian Qu}
  \state{Beijing Shi}
  \country{China}
}

\author{Zhijing Wu}
\affiliation{%
  \institution{Beijing Institute of Technology}
  \city{Haidian Qu}
  \state{Beijing Shi}
  \country{China}
}

\author{Xian-Ling Mao}
\affiliation{%
  \institution{Beijing Institute of Technology}
  \city{Haidian Qu}
  \state{Beijing Shi}
  \country{China}
}

\renewcommand{\shortauthors}{Sun et al.}


\begin{abstract}
    The rapid development of generative portrait models has raised growing concerns about privacy leakage and identity misuse. In particular, audio-driven 3D talking face generation can reconstruct a reusable 3D portrait of a target person from a monocular video and animate it with arbitrary speech, making realistic identity impersonation alarmingly practical. Existing proactive defenses mainly operate in the visual domain by injecting subtle perturbations into facial regions to disrupt identity acquisition. However, such perturbations often compromise visual quality due to the strong structural priors and social sensitivity of human faces, and are easily weakened by common real-world transformations such as resizing.
To overcome these limitations, we propose an imperceptible audio defense for audio-driven 3D talking face generation by shifting protection from the visual modality to the audio modality. Specifically, we exploit psychoacoustic masking to hide protective perturbations within perceptually masked frequency regions of the speech signal, thereby reducing perceptual distortion while suppressing reliable facial animation. Extensive experiments demonstrate that the proposed method effectively degrades 3D talking face generation while preserving favorable perceptual quality. These findings highlight psychoacoustically guided audio perturbations as a practical and promising direction for privacy-preserving portrait protection.
\end{abstract}


\begin{teaserfigure}
  \includegraphics[width=\textwidth]{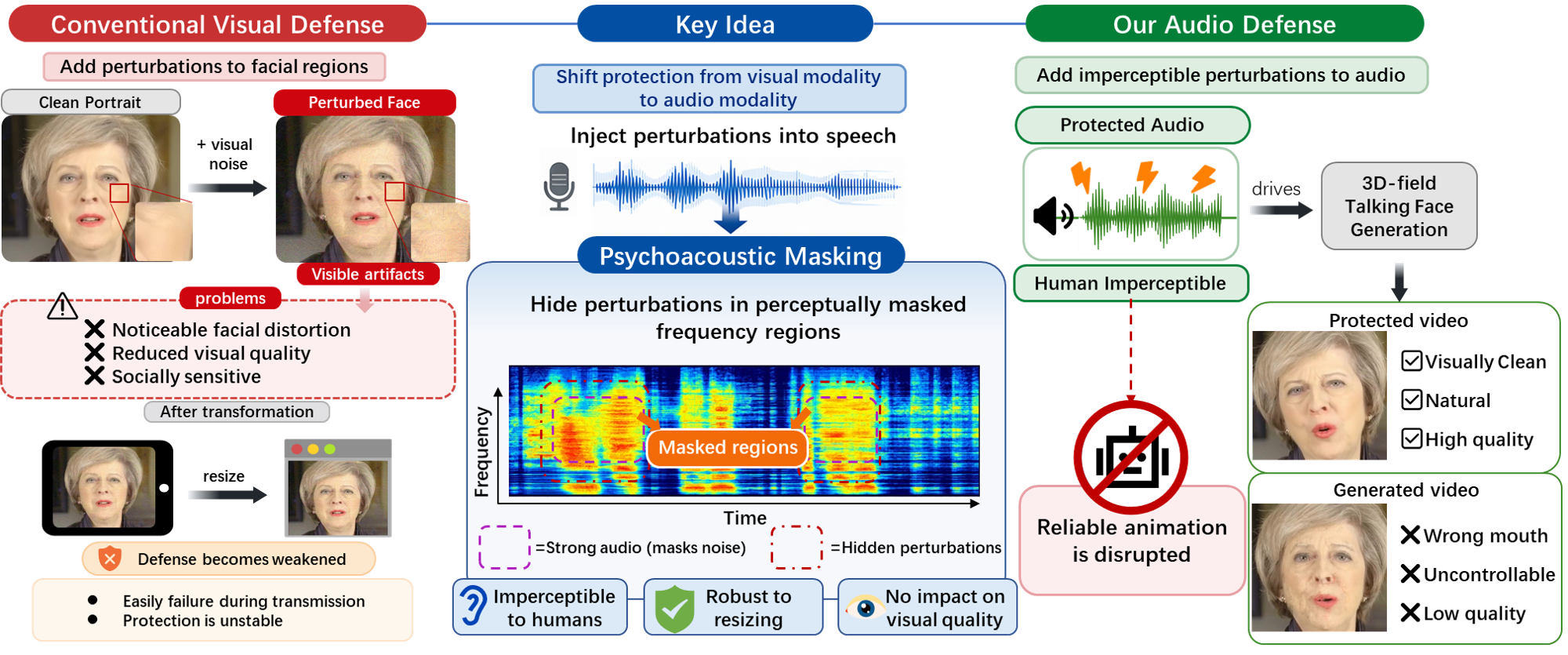}
  \caption{Comparison between conventional visual-domain defense and our psvchaacoustically quided audio-domain defense.
Visual perturbations may introduce noticeable artifacts and are vulnerable to resizing, while our method hides protective perturbations
in perceptuallv masked audio regions, preserving portrait appearance while suppressing reliable 3D-feild talking face generation.}
  \Description{Enjoying the baseball game from the third-base
  seats. Ichiro Suzuki preparing to bat.}
  \label{fig:teaser}
\end{teaserfigure}
\keywords{Talking Face Generation, Proactive Defense, Audio Adversarial Attack,}
\maketitle



\section{Introduction}

The recent surge of generative multimedia models has greatly expanded the capability of synthesizing realistic human-centric content, including portraits, speech, and talking videos \cite{rombach2022high, brooks2024instructpix2pix, ditto2025Ditto}. Among these advances, talking face generation has become a representative multimodal task because it tightly couples facial appearance with audio-driven motion generation \cite{chen2019_early_method_HCTFG, prajwal2020_WAV2LIP, Peng_2024_synctalk}. As a representative multimodal generation task, talking face generation tightly couples visual identity modeling with speech-driven motion synthesis, making privacy protection especially challenging in real-world multimedia sharing scenarios. In particular, audio-driven 3D talking face generation can recover a personalized 3D portrait from a monocular video and subsequently animate it with arbitrary speech, producing highly realistic and identity-consistent results \cite{li2023ernerf, Tang_2025_rad-nerf, Ye_2024_geneface, Peng_2024_synctalk, Li_2025_instag}. These capabilities open up promising opportunities for digital humans, virtual communication, and multimedia creation. \textbf{However,} the same capability also makes portrait misuse alarmingly practical. Once a victim's portrait information is acquired, the reconstructed 3D representation can be reused for scalable and near-real-time synthesis, enabling impersonation, fabricated speeches, and other forms of identity abuse \cite{verdoliva2020media, masood2023deepfakes}.

This privacy risk is especially concerning in today's online ecosystem, where user-generated videos are continuously shared across social platforms and can be easily harvested by malicious actors. Compared with conventional 2D portrait animation \cite{chen2019_early_method_HCTFG, prajwal2020_WAV2LIP}, audio-driven 3D talking face generation poses a more serious threat because it not only captures appearance details, but also reconstructs reusable three-dimensional identity cues that support high-fidelity reenactment under arbitrary speech input \cite{li2023ernerf, Peng_2024_synctalk, Li_2025_instag}. In practice, even a short publicly accessible video may already provide sufficient material for reconstructing a controllable digital portrait of the target person. Therefore, protecting portrait media before it is collected by such generation systems has become an urgent problem in multimedia security and privacy protection.

To mitigate these risks, recent proactive defense methods attempt to add subtle perturbations to media before they are scraped or reused by generative models. Most existing defenses are developed in the visual domain, where adversarial perturbations are injected into facial regions to interfere with identity extraction, landmark estimation, or downstream portrait reconstruction \cite{Salman_2023_advdm, Shan_2023_Mist, Salman_2023_PhotoGuard, FaceLock, faceShield}. This line of research is intuitive, since visual appearance is the most direct source for identity acquisition. Nevertheless, visual-domain protection faces inherent limitations when applied to portrait sharing scenarios. Human faces are highly structured and socially sensitive visual objects, and even slight distortions in facial regions may noticeably reduce visual quality, damage user experience, or introduce unnatural artifacts. More importantly, visual perturbations are often fragile under common real-world transformations such as resizing, resampling, and compression, which are almost unavoidable during online transmission and platform processing \cite{DFRAP}. As a result, their protection effectiveness can be significantly weakened before the media is actually used by an attacker.

These limitations motivate us to rethink \emph{where} protection should be imposed in audio-driven 3D talking face generation. Instead of continuing to perturb the visually sensitive facial region, we ask whether the defense can be shifted to the audio modality. This perspective is particularly attractive because the driving audio plays a fundamental role in controlling lip motion and facial dynamics, making it a natural intervention point for disrupting high-quality synthesis. Meanwhile, unlike the visual domain, the auditory domain provides a more explicit perceptual principle for imperceptibility, namely psychoacoustic masking \cite{qin2019imperceptible, schonherr2019adversarial}. According to this principle, weak sound components can become imperceptible in the presence of stronger neighboring frequencies, which provides a principled way to hide protective perturbations while preserving perceptual quality.

Based on this insight, we propose an imperceptible audio defense for audio-driven 3D talking face generation. Rather than directly perturbing facial pixels, our method injects protective perturbations into perceptually masked frequency regions of the speech signal under psychoacoustic guidance. In this way, the perturbations remain difficult for human listeners to perceive, while still interfering with the generation pipeline and suppressing reliable facial animation. By shifting protection from the visual modality to the audio modality, our approach avoids directly damaging portrait appearance and offers a more user-friendly solution for privacy-preserving portrait sharing.

Extensive experiments demonstrate that the proposed method effectively degrades the performance of audio-driven 3D talking face generation while preserving favorable perceptual quality. The results suggest that psychoacoustically guided audio perturbations provide a practical and promising direction for proactive portrait protection, especially in scenarios where visual fidelity is critical. Overall, this work offers a new modality-level perspective on defending against portrait generation systems and highlights the importance of incorporating perceptual principles into deployable multimedia privacy protection methods.

Our main contributions are summarized as follows:
\begin{itemize}
    \item We revisit proactive defense for talking face generation from a new modality perspective and reveal the limitations of existing visual-domain protection methods in terms of perceptual quality and robustness under common real-world transformations.
    \item We propose an imperceptible audio defense for audio-driven 3D talking face generation by leveraging psychoacoustic masking to conceal protective perturbations in perceptually masked frequency regions.
    \item We demonstrate through extensive experiments that the proposed method effectively suppresses 3D talking face generation while maintaining favorable perceptual quality, showing the practicality of audio-based protection for privacy-preserving portrait sharing.
\end{itemize}

\section{Related Works}

\subsection{Audio-driven Talking Face Generation}

Audio-driven talking face generation (TFG) aims to synthesize realistic talking portraits by animating facial motions according to speech signals. Early methods mainly relied on 2D image generation or warping-based pipelines \cite{chen2019_early_method_HCTFG, chen2018_lipGan, prajwal2020_WAV2LIP}, which achieved plausible lip synchronization but often struggled to model natural head movement, fine-grained facial dynamics, and multi-view consistency. 

Recent advances in 3D representations have substantially improved the realism and controllability of TFG. In particular, Neural Radiance Fields (NeRF) \cite{Athar_2022_RigNeRF, Bi_2024_NeRF-AD, guo2021adnerf, Tang_2025_rad-nerf, li2023ernerf, Ye_2024_geneface, Peng_2024_synctalk} and 3D Gaussian Splatting (3DGS) \cite{cho2024gaussiantalker, Li_2024_talkingaussian, Li_2025_instag} enable the reconstruction of subject-specific 3D portraits from monocular reference videos, leading to stronger identity consistency, more accurate lip synchronization, and better view-consistent rendering. Compared with generalized one-shot portrait animation methods \cite{ditto2025Ditto, ye2024real3dportrait}, these personalized 3D-field approaches are particularly powerful because they can recover reusable identity-aware 3D representations and support highly realistic speech-driven reenactment.

However, this strong generation capability also introduces significant privacy risks. Once a target person’s reference video is collected, current 3D-field TFG models can reconstruct a controllable digital portrait and synthesize realistic videos under arbitrary speech input. This makes them especially concerning in public multimedia sharing scenarios, where user portraits and voices can be easily harvested and reused. In this work, we focus on defending against such \emph{audio-driven 3D-field TFG systems}, which pose a more serious threat than conventional 2D talking face generation due to their stronger realism, controllability, and reusability.

\subsection{Talking Face Defense}

With the rapid progress of portrait generation models, recent studies have begun to explore \emph{proactive defenses} against talking face generation and related portrait manipulation systems. Most existing methods are developed in the \emph{visual domain}, where protective perturbations are injected into facial regions to disrupt identity extraction, facial landmark estimation, reconstruction, or downstream editing and animation models \cite{Salman_2023_advdm, Shan_2023_Mist, Salman_2023_PhotoGuard, faceShield, FaceLock}. This line of work is intuitive because the face image is the most direct source for identity modeling and visual synthesis.

Despite their effectiveness, visual-domain defenses face two major limitations in the context of talking face protection. First, human faces are highly structured and socially sensitive visual objects, so even subtle perturbations may noticeably reduce visual quality or introduce undesirable facial artifacts. Second, such perturbations are often fragile under common real-world transformations, including resizing, resampling, and platform-side compression, which are almost unavoidable during online media sharing and redistribution. These issues are especially problematic for 3D-field TFG, where accurate visual priors are often reconstructed through long video preprocessing pipelines and may be partially restored or purified during fitting.

Compared with prior work, our goal is not to further optimize visual perturbations, but to revisit \emph{where} protection should be imposed. We argue that, for audio-driven talking face generation, the \emph{driving audio itself} provides a more natural and practical intervention point. Instead of perturbing visually sensitive facial regions, we shift protection to the audio modality and study how imperceptible audio perturbations can suppress reliable facial animation while preserving portrait appearance. To the best of our knowledge, this perspective remains largely underexplored in proactive defense for talking face generation, especially for modern 3D-field TFG systems.

\subsection{Voice Clone Defense}

Our work is also related to recent efforts on \emph{voice clone defense} and adversarial protection for speech generation systems. Prior studies have shown that speech models, including automatic speech recognition (ASR), speaker verification, and voice cloning systems, are vulnerable to carefully designed perturbations in the audio domain \cite{carlini2018audio, qin2019imperceptible, schonherr2019adversarial, neekhara2019universal, yakura2019robust}. To improve perceptual stealth, several works further incorporate psychoacoustic principles and human auditory masking models to constrain perturbations below perceptual thresholds \cite{qin2019imperceptible, schonherr2019adversarial}. These studies demonstrate that the audio modality provides a principled path to achieving both attack effectiveness and human imperceptibility.

However, defending against audio-driven talking face generation is fundamentally different from defending against pure speech or speaker models. In our setting, the audio signal is not the final target itself, but a \emph{cross-modal control signal} that drives facial geometry and lip motion. Therefore, the objective is not merely to alter linguistic content or speaker identity, but to disrupt the \emph{audio-to-geometry mapping} required for realistic facial animation. This introduces a new challenge: the perturbation must remain imperceptible to listeners while still being strong enough to interfere with the downstream visual generation process. Our work bridges this gap by introducing a psychoacoustically guided audio defense specifically designed for 3D talking face generation, connecting voice-side imperceptible perturbation design with portrait privacy protection.

\section{Preliminaries}
\subsection{Audio-Driven Talking Face Generation in 3D Fields}
\begin{figure}[t]
  \centering
  \includegraphics[width=\linewidth]{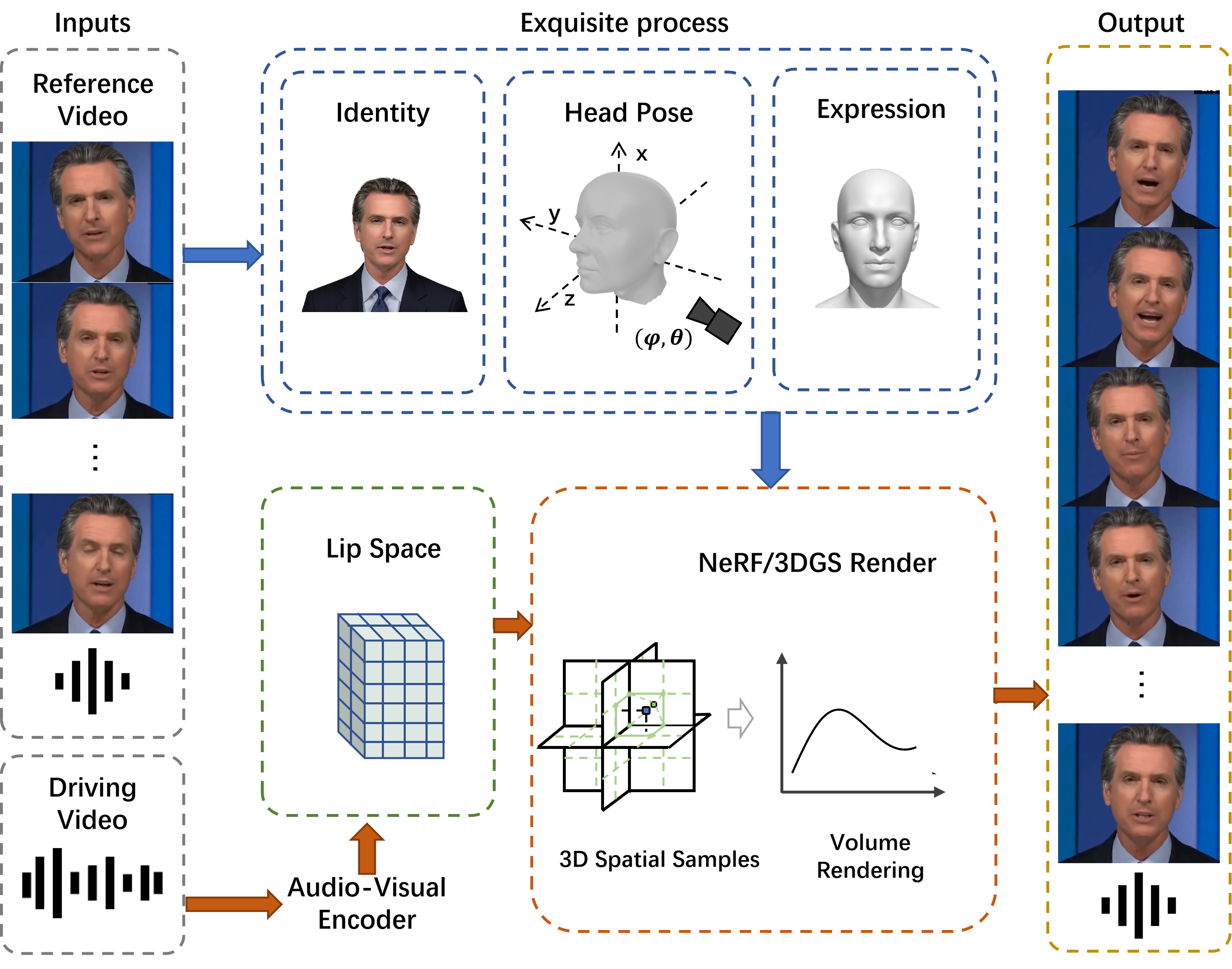} 
  \caption{Overview of the 3D-field TFG pipeline. The reference video is disentangled into identity, pose, and expression priors. Driving audio modulates 3D spatial samples via a latent lip space. Blue arrows indicate training-only operations (e.g., identity/pose extraction), while red arrows denote operations common to both training and inference (e.g., audio-visual encoding and neural rendering).}
  \label{fig:tfg_pipeline}
\end{figure}
Unlike 2D-based talking head synthesis that operates on image pixels, 3D-field TFG models represent the human head as a continuous or discrete dynamic 3D structure. This approach ensures multi-view consistency and realistic geometric deformation. The common pipeline involves mapping acoustic features to a 3D spatial representation, which is then rendered into 2D images.

\noindent \textbf{Generalized 3D Scene Representations.} 
A 3D field $\mathcal{F}$ characterizes the physical properties (e.g., density $\sigma$, color $\mathbf{c}$, or covariance $\Sigma$) of any point $\mathbf{x} \in \mathbb{R}^3$. Recent TFG frameworks employ various efficient 3D representations:

\begin{itemize}
    \item {\texttt{Volumetric Fields}}: (NeRF-based) The scene is represented as a continuous function $f: (\mathbf{x}, \mathbf{d}) \to (\sigma, \mathbf{c})$. To accelerate rendering, Tri-planes ($\mathcal{P}_{xy}, \mathcal{P}_{xz}, \mathcal{P}_{yz}$) are often employed, where features at coordinate $\mathbf{x}$ are aggregated from orthogonal projections:
    \begin{equation}
        \boldsymbol{f}_{\mathbf{x}} = \operatorname{Agg}_{i \in \{xy, xz, yz\}} \left( \mathcal{I}(\mathcal{P}_{i}, \pi_{i}(\mathbf{x})) \right)
    \end{equation}

    \item {\texttt{Point-based Fields}}: (3DGS-based) The scene is explicitly modeled by a collection of 3D Gaussians, each defined by its center $\boldsymbol{\mu}$, opacity $\alpha$, and covariance matrix $\boldsymbol{\Sigma}$. Audio features drive the facial motion by predicting residual offsets $\Delta\boldsymbol{\mu}$ or modulating a canonical deformation field.

    \item {\texttt{Hybrid Fields}}: These models leverage 3D Morphable Models (3DMM) to provide explicit structural priors ($\boldsymbol{\alpha}, \boldsymbol{\beta}, \boldsymbol{\delta}$), integrating parametric facial geometry with neural feature volumes to achieve both high-fidelity rendering and precise expression control.
\end{itemize}

\medskip

\noindent \textbf{Audio-to-Geometry Mapping.} 
The fundamental mechanism of a 3D-field TFG model $G$ is the modulation of the 3D field $\mathcal{F}$ by an audio embedding $\boldsymbol{a}$. Let $\Phi$ be the transformation function that maps the canonical identity to a dynamic state:
\begin{equation}
    \mathcal{F}_{sync} = \Phi(\mathcal{F}_{static}, \boldsymbol{a})
\end{equation}
\noindent where $\mathcal{F}_{static}$ represents the static spatial prior of the speaker, and $\mathcal{F}_{sync}$ is the audio-synced dynamic field. 



\subsection{Principles of Psychoacoustic Masking}
To achieve imperceptible protection, we must ensure that the adversarial noise added to the audio remains beneath the detection threshold of the human auditory system. This is governed by the Psychoacoustic Masking Effect.

\noindent \textbf{Frequency Masking and Critical Bands.} The human ear processes sound through a non-linear filter bank on the cochlea, known as Critical Bands. A strong sound at a specific frequency (the masker) will raise the hearing threshold of neighboring frequencies (the maskee). This masking power is more effectively modeled on the Bark Scale ($z$):
\begin{equation}
z = 13 \arctan(0.00076 f) + 3.5 \arctan((f / 7500)^2)
\end{equation}
Within this scale, if the energy of our adversarial perturbation $\delta$ stays within the "masking umbrella" created by the original speech signal $x$, the human brain will filter out $\delta$ as redundant information.

\noindent \textbf{Absolute Threshold of Hearing (ATH).} The Absolute Threshold of Hearing (ATH) represents the baseline sensitivity of the human ear in a noiseless environment. It is defined by the non-linear function $T_q(f)$:
\begin{equation}
T_q(f) = 3.64(f/1000)^{-0.8} - 6.5e^{-0.6(f/1000-3.3)^2} + 10^{-3}(f/1000)^4 \text{ (dB)}
\end{equation}
This threshold indicates that the ear is most sensitive between $2\text{--}5$ kHz (the range of human speech) and significantly less sensitive at very low or very high frequencies. Any protective noise falling below $T_q(f)$ is physically inaudible.

\begin{figure}[t]
  \centering
  \includegraphics[width=\linewidth]{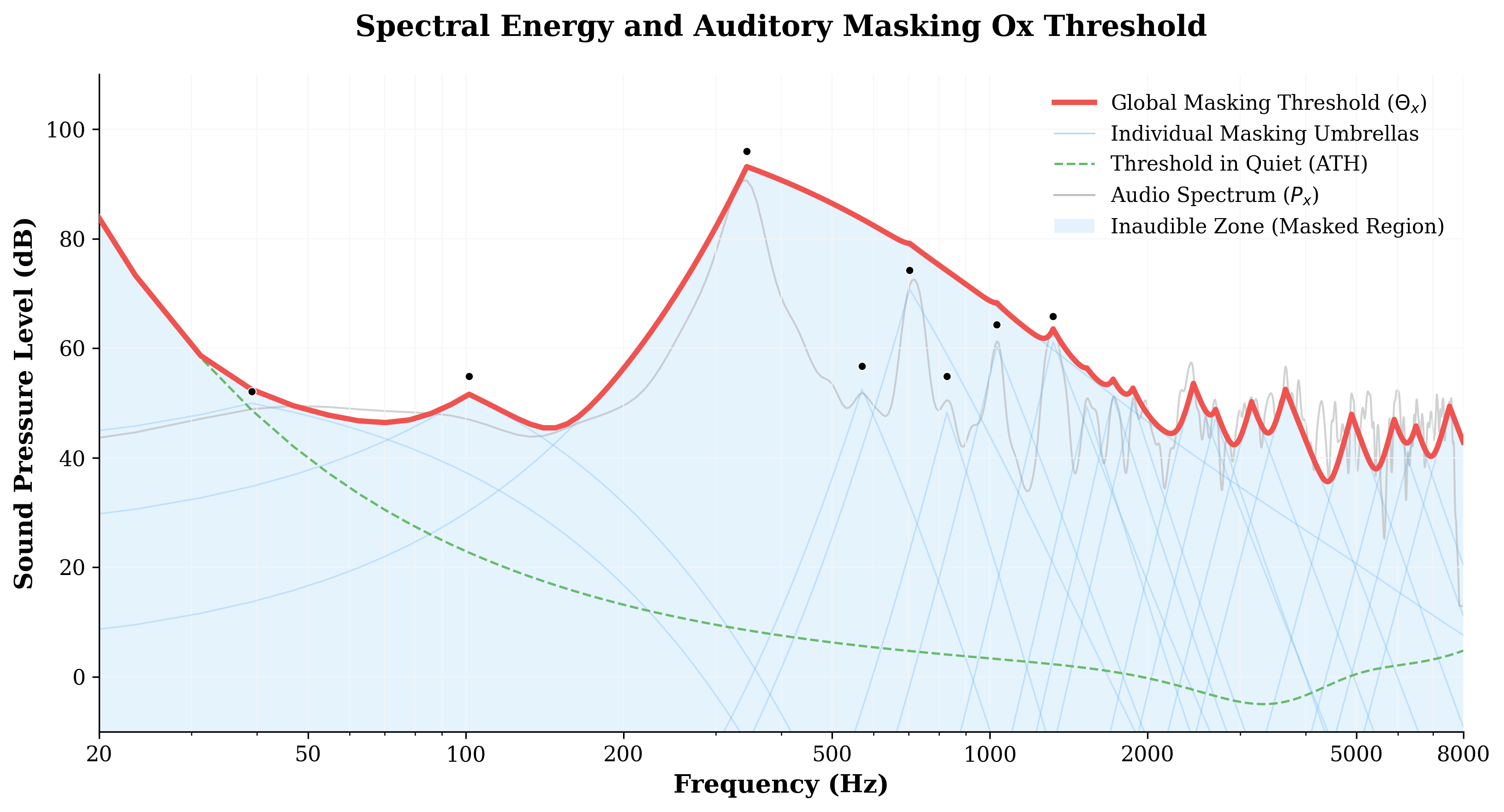}
  \caption{Visualization of the frequency masking mechanism for a speech segment. The clean audio spectrum $P_x$ (light gray) creates multiple individual masking umbrellas (light blue) centered at dominant tonal maskers (black dots). The Global Masking Threshold $\Theta_x$ (red thick line) represents the aggregated perceptual boundary. Any adversarial perturbation $\delta$ residing within the Inaudible Zone (blue shaded area) is rendered imperceptible to the human ear, as it remains below the frequency-dependent sensitivity defined by the Absolute Threshold of Hearing (ATH, green dashed line) and the simultaneous masking effects.}
  \label{fig:masking_vis}
\end{figure}

\noindent \textbf{Global Masking Threshold $\theta(k)$.} The Global Masking Threshold is the combination of the ATH and the simultaneous masking effects produced by all components of the speech signal. The process of computing $\theta(k)$ involves: (i) \textit{Decomposition}: Identifying tonal and non-tonal maskers in the PSD of $x$; (ii) \textit{Spreading}: Applying a spreading function $SF(z_i, z_j)$ to model the frequency-dependent decay; and (iii) \textit{Aggregation}: Summing individual thresholds and the ATH in the intensity domain:
\begin{equation}
    \theta(j) = 10 \log_{10} \left( 10^{T_q(j)/10} + \sum_{i=1}^{M} 10^{T_i(j)/10} \right)
\end{equation}
As illustrated in Fig. \ref{fig:masking_vis}, our strategy embeds the protective perturbations $\delta$ within the resulting "Inaudible Zone" (shaded in blue). By enforcing the constraint $\text{PSD}(\delta) \leq \theta(j)$, we guarantee that the defense remains transparent to the user, effectively hiding the semantic conflict within the perceptual shadows of the speech signal.

\subsection{Threat Model and Defense Objective}
\noindent \textbf{Threat Model.} 
In this study, we consider a security scenario involving a victim user who shares their original audio $x$ on public social media platforms. An unauthorized adversary, acting as a deepfake creator, aims to harvest this audio to synthesize a highly deceptive talking-face video using a 3D-field TFG model $M$. 

\begin{itemize}
    \item \texttt{Adversarial Intent:} The adversary seeks to exploit the victim’s acoustic identity to drive a digital avatar, ensuring that the synthesized lip movements are perfectly synchronized with the victim's speech to facilitate social engineering or misinformation.
    \item \texttt{Black-box Assumptions:} We assume a stringent black-box setting where the adversary has no prior knowledge of our defensive mechanism, its parameters, or the underlying psychoacoustic constraints. This forces the adversary to rely solely on the perceived quality of the audio for synthesis.
    \item \texttt{Channel Robustness:} Considering the practical transmission of audio over the Internet, we assume the adversary operates on a version of $x$ that has undergone standard lossy compression (e.g., MP3 or AAC transcoding) and resampling. Our threat model stipulates that a successful defense must remain effective even after such signal distortions, which are typical of social media pipelines.
\end{itemize}

\noindent \textbf{Defense Objective.}
We seek a perturbation $\delta$ such that $x_{adv} = x + \delta$ achieves: (1) \textbf{Visual Targeted Disruption}, where the cross-modal conflict causes geometric collapse in the mouth region; and (2) \textbf{Acoustic Imperceptibility}, satisfying $\mathcal{D}_{perc}(x, x_{adv}) < \epsilon$ based on $\theta(k)$.

\section{Methodology}

\begin{figure*}[t]
    \centering
    \includegraphics[width=1.0\linewidth]{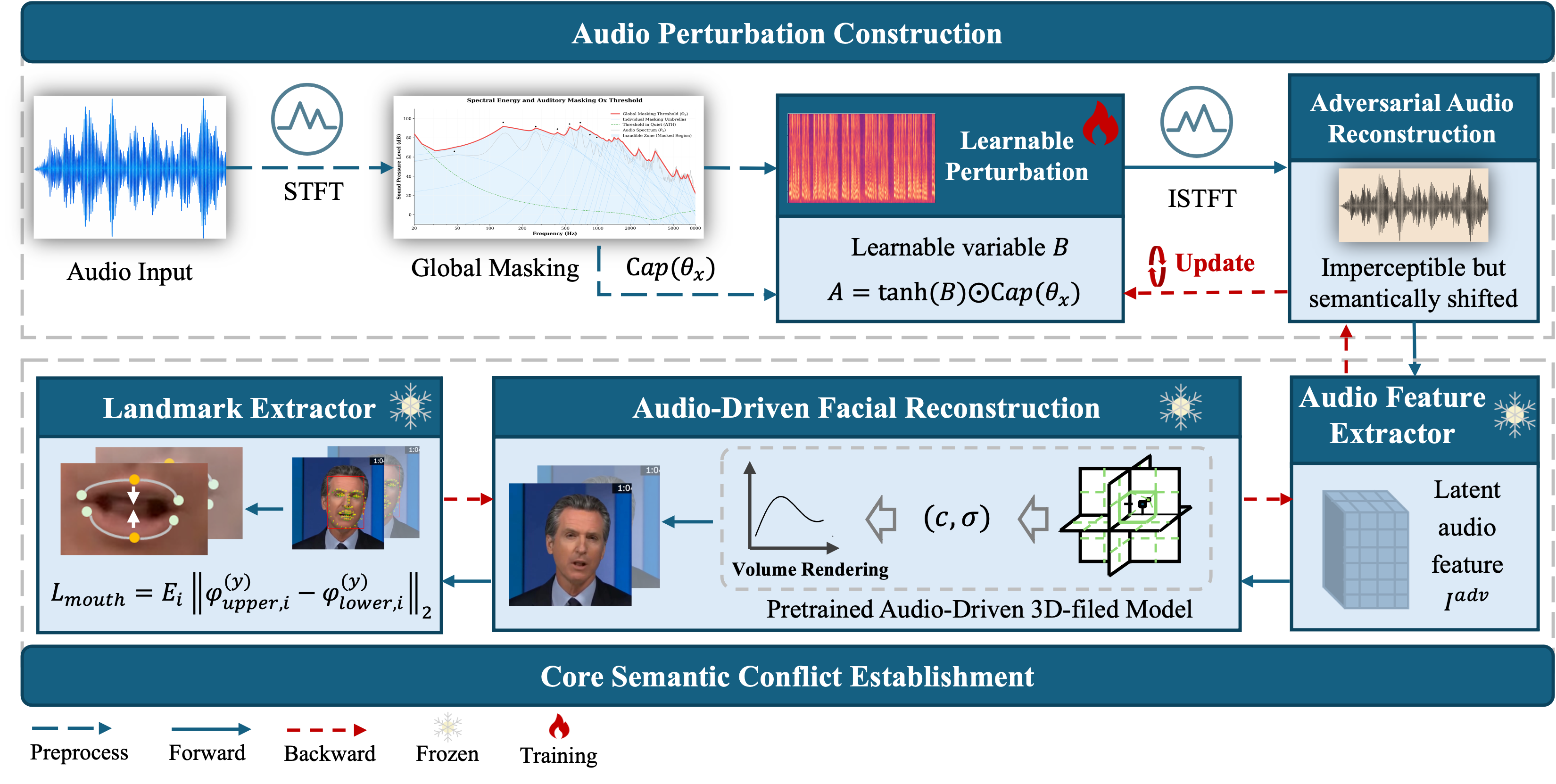}
    \caption{Overview of the proposed visually lossless proactive defense framework. The upper branch constructs a psychoacoustically bounded frequency-domain perturbation and reconstructs an adversarial waveform that is imperceptible to human listeners but semantically shifted for the talking-face generator. The lower branch establishes cross-modal semantic conflict by feeding the perturbed audio into a frozen audio feature extractor and a frozen pretrained audio-driven 3D-field renderer, while a mouth landmark loss encourages collapsed mouth geometry in the rendered results. By optimizing the perturbation under psychoacoustic constraints, our method disrupts the audio-to-geometry mapping without introducing visible artifacts into the shared visual content.}
    \label{fig:method}
\end{figure*}

Our goal is to construct an \emph{imperceptible} audio perturbation that preserves the perceptual quality of the shared media while preventing it from serving as a reliable driving signal for personalized 3D talking face generation. As illustrated in Fig.~\ref{fig:method}, the proposed framework consists of two tightly coupled stages: \textbf{Audio Perturbation Construction} and \textbf{Core Semantic Conflict Establishment}. The first stage generates a psychoacoustically bounded perturbation in the frequency domain, and the second stage feeds the perturbed audio into a pretrained audio-driven 3D-field talking face model to establish a persistent mismatch between audio semantics and mouth geometry. The entire process is optimized end-to-end through the asynchronous dual-loop strategy summarized in Algorithm~\ref{alg:3dmm_attack}.

\subsection{Framework Overview}

The key idea of our method is to shift the defense from the visual domain to the audio domain. Instead of directly injecting perturbations into facial pixels, we optimize a frequency-domain perturbation on the driving speech such that the resulting audio remains perceptually natural to humans, but induces misleading semantic cues for the downstream talking-face generator. Once the perturbed audio is used to drive facial reconstruction, the generated mouth motion becomes structurally inconsistent with the expected speech articulation, thereby breaking the learned audio-to-lip correspondence required by personalized 3D talking face generation.

As shown in Fig.~\ref{fig:method}, the upper branch first transforms the clean waveform into the time-frequency domain, computes a psychoacoustic capacity bound from the masking threshold, and optimizes a learnable perturbation variable within this admissible region. The perturbed spectrum is then converted back to the time domain to obtain the adversarial waveform. The lower branch feeds the adversarial waveform into a frozen audio feature extractor and a frozen pretrained audio-driven 3D-field renderer. A differentiable landmark extractor is finally applied to the rendered frames, and the resulting mouth closure loss is back-propagated to update the perturbation. In this way, the optimized perturbation does not visibly alter the shared video, yet it poisons the semantic driving signal used by the talking-face model.

\subsection{Audio Perturbation Construction}

We first construct the perturbation in the frequency domain, since psychoacoustic masking constraints are naturally defined over time-frequency bins. Given a clean waveform $x \in [-1,1]^T$, we compute its short-time Fourier transform:
\begin{equation}
S_x = \mathrm{STFT}(x),
\end{equation}
where $S_x \in \mathbb{C}^{F \times T_s}$ denotes the complex spectrum, with $F$ frequency bins and $T_s$ temporal frames. We further decompose it into magnitude and phase, and define the phase template as
\begin{equation}
P = \exp\!\left(j \angle S_x\right).
\end{equation}

Based on the clean spectrum, we estimate the psychoacoustic masking threshold map $\Theta_x$, which specifies the maximum imperceptible perturbation energy at each time-frequency location. From $\Theta_x$, we derive a perturbation capacity map $C$, which provides an upper bound for the perturbation amplitude. To ensure that the optimization always remains within the admissible psychoacoustic region, we introduce a learnable variable $B$ and parameterize the perturbation amplitude as
\begin{equation}
A = \tanh(B) \odot C,
\label{eq:A_from_B}
\end{equation}
where $\odot$ denotes element-wise multiplication. The $\tanh(\cdot)$ mapping constrains each element of $B$ to $(-1,1)$, so that the resulting perturbation amplitude $A$ never exceeds the psychoacoustic budget defined by $C$.

We then construct the complex perturbation
\begin{equation}
\Delta = A \odot P,
\end{equation}
and add it to the clean spectrum to obtain the adversarial spectrum
\begin{equation}
S_{adv} = S_x + \Delta.
\end{equation}
Finally, the adversarial waveform is reconstructed by inverse STFT:
\begin{equation}
x^{adv} = \mathrm{ISTFT}(S_{adv}).
\end{equation}

This design corresponds to the upper branch of Fig.~\ref{fig:method}. It also matches the first several lines of Algorithm~\ref{alg:3dmm_attack}: we compute $S_x$, the phase template $P$, the threshold map $\Theta_x$, the perturbation bound $C$, and optimize the learnable variable $B$ through iterative updates.

\subsection{Core Semantic Conflict Establishment}

After obtaining the adversarial waveform $x^{adv}$, we use it to drive a frozen audio-driven 3D-field talking face model. This stage corresponds to the lower branch of Fig.~\ref{fig:method}. Specifically, the adversarial waveform is first mapped into latent audio features:
\begin{equation}
F^{adv} = \mathcal{E}(x^{adv}),
\end{equation}
where $\mathcal{E}$ denotes the frozen audio feature extractor. These features are then fed into a pretrained audio-driven renderer $\mathcal{R}$ together with the visual conditioning of the target identity to synthesize talking-face frames:
\begin{equation}
\hat{I}_i = \mathcal{R}(d_i, a_i), \quad i = 1,\dots,N,
\end{equation}
where $d_i$ denotes the visual or camera-related condition of frame $i$, and $a_i$ is the audio condition derived from $F^{adv}$.

The central objective of our method is not to suppress the visual content itself, but to establish a semantic conflict between the perturbed audio cues and the expected mouth articulation. To quantify this conflict, we apply a differentiable landmark detector $\mathcal{D}$ to each rendered frame and obtain the mouth landmarks:
\begin{equation}
L_i = \mathcal{D}(\hat{I}_i).
\end{equation}
Let $\phi_{upper,i}^{(y)}$ and $\phi_{lower,i}^{(y)}$ denote the vertical coordinates of the selected upper and lower inner-lip landmarks in frame $i$, respectively. We define the mouth deconstruction loss as
\begin{equation}
L_{mouth}
=
\frac{1}{N}
\sum_{i=1}^{N}
\left\|
\phi_{upper,i}^{(y)}-\phi_{lower,i}^{(y)}
\right\|_2.
\label{eq:L_mouth}
\end{equation}
Minimizing $L_{mouth}$ encourages the rendered mouth to collapse toward a closed state, even when the original clean speech should correspond to open-mouth articulations. This mismatch forces the model into an inconsistent audio-to-geometry mapping regime, thereby weakening its ability to learn stable and accurate lip motion patterns from the protected data.

This process is exactly reflected in the middle part of Algorithm~\ref{alg:3dmm_attack}: after generating $x^{adv}$ and extracting $F^{adv}$, the renderer synthesizes frames one by one, the landmark detector analyzes each frame, and the mouth-related loss is accumulated over all rendered frames.

\subsection{Optimization Objectives}

To jointly enforce attack effectiveness, perceptual naturalness, and spectral smoothness, we optimize the perturbation using four complementary objectives.

\paragraph{Mouth deconstruction loss.}
The primary attack objective is the mouth closure loss:
\begin{equation}
L_{atk} = L_{mouth}.
\end{equation}
It directly measures whether the perturbed audio induces collapsed or semantically incorrect mouth motion in the rendered talking-face outputs.

\paragraph{Spectral energy regularization.}
To avoid trivial solutions such as suppressing the signal energy, we constrain the magnitude of the adversarial spectrum to remain close to that of the clean spectrum:
\begin{equation}
L_{eng}
=
\mathbb{E}_{k,t}
\left(
|S_{adv}(k,t)| - |S_x(k,t)|
\right)^2.
\label{eq:L_eng}
\end{equation}
This term encourages the defense to rely on semantic manipulation rather than brute-force signal destruction.

\paragraph{Psychoacoustic masking loss.}
Although the parameterization in Eq.~(\ref{eq:A_from_B}) already limits the perturbation within a bounded region, we further penalize any residual violation of the masking threshold using
\begin{equation}
L_{mask}
=
\mathbb{E}_{k,t}
\left[
\max\!\left(
\bar{p}_{\delta}(k,t)-\Theta_x(k,t),\, 0
\right)
\right],
\label{eq:L_mask}
\end{equation}
where $\bar{p}_{\delta}(k,t)$ denotes the normalized perturbation power spectral density. This term encourages the perturbation to stay in the auditory masking shadow of the clean audio.

\paragraph{Temporal-frequency smoothness.}
To prevent isolated spikes and unnatural spectral artifacts, we regularize the perturbation amplitude with first- and second-order smoothness constraints:
\begin{equation}
L_{smooth}
=
\|\nabla_t A\|_1 + \|\nabla_f A\|_1
+
\|\nabla_t^2 A\|_1 + \|\nabla_f^2 A\|_1,
\label{eq:L_smooth}
\end{equation}
where $\nabla_t$ and $\nabla_f$ denote temporal and frequency derivatives, respectively. This regularization makes the perturbation more consistent with the natural continuity of speech spectra.

\subsection{Asynchronous Dual-Loop Optimization}

The full optimization procedure is summarized in Algorithm~\ref{alg:3dmm_attack}. Since the attack objective and the perceptual objective have inherently different priorities, we adopt an asynchronous dual-loop strategy instead of optimizing all losses with a single static weighting scheme.

\paragraph{Primary optimization loop.}
In each iteration, we first construct the perturbation amplitude $A$, reconstruct the adversarial waveform $x^{adv}$, and feed it into the audio-driven talking-face pipeline. The rendered frames are then evaluated by the landmark detector to compute the mouth deconstruction loss $L_{atk}$. Together with the spectral energy loss and smoothness regularization, the main objective is defined as
\begin{equation}
L_{\mathrm{main}}
=
L_{atk}
+
\lambda_{eng} L_{eng}
+
\lambda_{smooth} L_{smooth}.
\label{eq:L_main}
\end{equation}
This objective is responsible for pushing the perturbation toward an effective semantic attack direction while maintaining reasonable spectral behavior.

\paragraph{Threshold activation.}
Once the attack objective reaches a predefined level, i.e., $L_{atk} \le \tau$, we mark the current perturbation as an effective candidate and activate the secondary refinement branch. Intuitively, this means the semantic conflict has already been established strongly enough to mislead the talking-face model.

\paragraph{Secondary refinement loop.}
After the attack threshold is met, we further compress the perturbation under psychoacoustic constraints, as long as the attack effect does not significantly deteriorate. The refinement objective is
\begin{equation}
L_{\mathrm{psy}}
=
\lambda_{mask} L_{mask}
+
\lambda_{smooth} L_{smooth}
+
\lambda_{reg}\|B\|_1.
\label{eq:L_psy}
\end{equation}
This stage corresponds to the red dashed update path in Fig.~\ref{fig:method} and to the final conditional branch of Algorithm~\ref{alg:3dmm_attack}. It reduces perceptual saliency and redundant perturbation energy while preserving attack effectiveness.

\paragraph{Discussion.}
The resulting optimization procedure has a clear interpretation. The upper branch of Fig.~\ref{fig:method} ensures that the perturbation is constructed in a psychoacoustically safe region, yielding an adversarial waveform that is \emph{imperceptible but semantically shifted}. The lower branch uses this shifted audio signal to drive a frozen talking-face generator and explicitly enforces incorrect mouth dynamics through landmark-based supervision. Therefore, unlike prior visual-domain defenses that sacrifice facial quality by perturbing pixels directly, our method attacks the cross-modal driving signal itself and achieves visually lossless protection against personalized 3D talking face generation.

\section{Experiments}
\begin{figure*}[t!]
    \centering
    \includegraphics[width=1.0\linewidth]{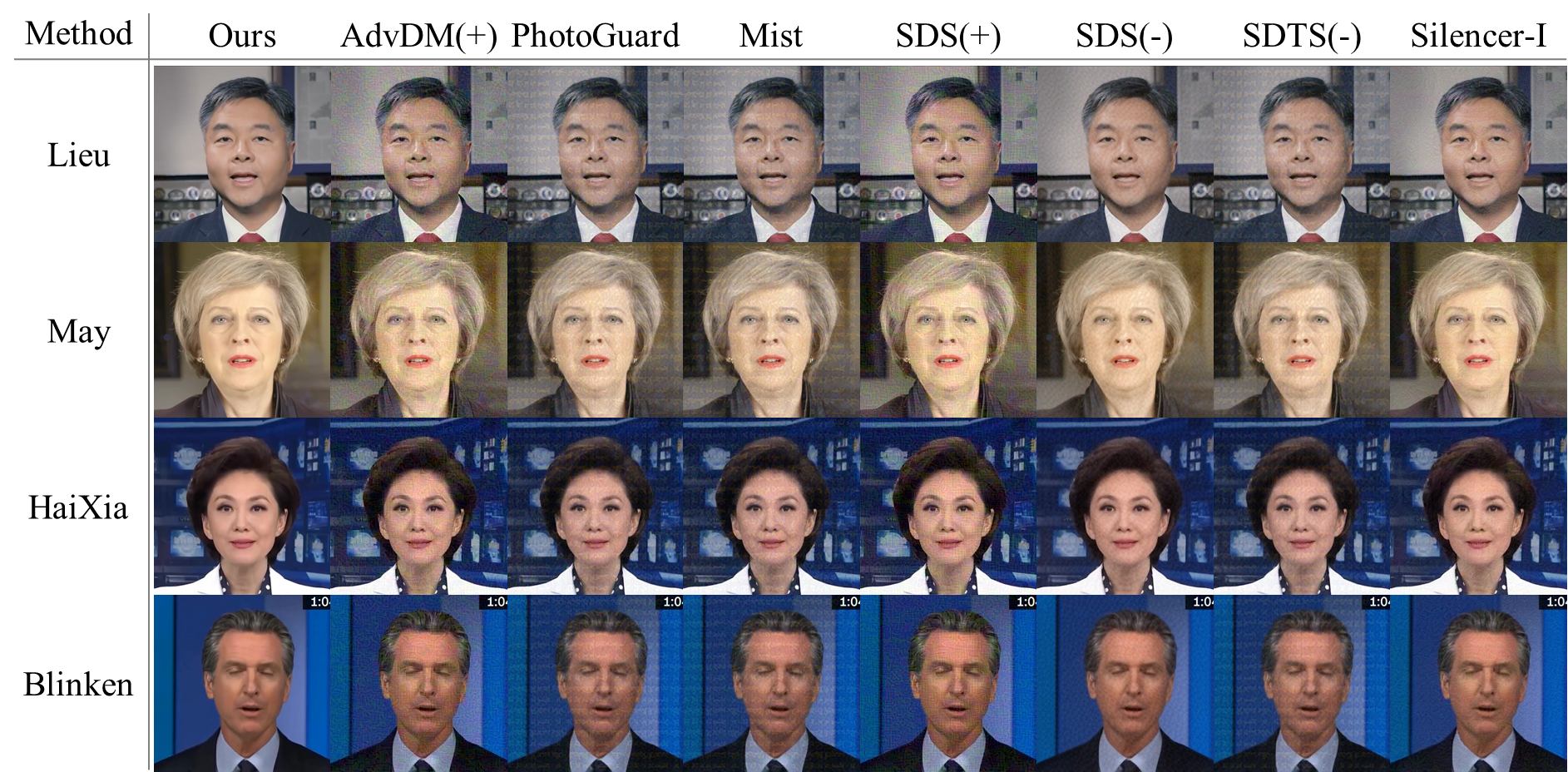}
    \caption{Qualitative comparison with visual-domain defense methods. From left to right, each row shows the clean reference result and the outputs under different defenses. Existing visual perturbation methods introduce noticeable artifacts in facial regions, such as grainy noise, texture corruption, and color distortion. In contrast, our method maintains visually lossless appearance while still preventing accurate speech-driven talking-face generation.}
    \label{fig:quality}
\end{figure*}
Due to space limitations, we present the main quantitative and representative qualitative results in the paper, and provide additional visual comparisons in the supplementary materials for a more comprehensive evaluation. We strongly recommend watching the supplementary video.

\subsection{Experimental Settings and Baselines}

To ensure a fair and practical evaluation, we collect the demonstration videos used in prior works and construct a subset of the HDTF \cite{zhang2021hdtf} dataset consisting of 11 identities. This setup balances computational cost and evaluation fairness, while covering diverse speaking styles, facial structures, and mouth motion patterns.

We compare our method against two representative baselines: \textit{antifake} \cite{yu2023antifake}, a visual-domain proactive defense method, and \textit{silencer} \cite{gan2025silence}, an audio-removal based defense strategy. These baselines represent two fundamentally different protection paradigms: adding perturbations to the visual signal versus directly suppressing the driving audio.

For evaluation, we adopt both generation-quality and perceptual-audio metrics. We denote metrics with the prefix ``M'' to indicate that they are computed over the mouth region, which is most relevant to speech-driven lip dynamics.

Specifically, we use \textbf{M-LMD}, \textbf{M-SSIM}, and \textbf{M-PSNR} to evaluate the impact of different defenses on generated talking-face videos, where all metrics are computed with respect to the outputs produced from clean reference videos. \textbf{M-LMD} measures the distance between mouth landmarks and reflects lip motion consistency, with lower values indicating better preservation of natural articulation. \textbf{M-SSIM} and \textbf{M-PSNR} measure structural similarity and pixel-level fidelity within the mouth region, respectively, where higher values indicate better visual quality.

In addition, we adopt \textbf{CDPAM} \cite{cdpam} to measure the perceptual distance between the defended audio and the clean audio, where lower values indicate higher perceptual similarity.

\subsection{Quantitative Results}

\begin{table}[t]
\centering
\caption{Comparison of different defense methods on lip-motion disruption, deviation from clean-driven generation, and perceptual audio quality.}
\label{tab:metric_comparison}
\begin{tabular}{lcccc}
\toprule
\textbf{Methods} & \textbf{M-LMD}$\uparrow$ & \textbf{M-SSIM}$\downarrow$ & \textbf{M-PSNR}$\downarrow$ & \textbf{CDPAM}$\downarrow$ \\
\midrule
antifake & 3.32034 & 0.83584 & 28.12566 & 0.457559 \\
silencer & 10.0028 & 0.5909 & 28.16 & 0 \\
ours     & 3.23093 & 0.84236 & 28.34491 & 0.116151 \\
\bottomrule
\end{tabular}
\end{table}

Table~\ref{tab:metric_comparison} reports the quantitative comparison across different defense methods. Since the objective of proactive defense is to prevent the protected media from serving as a reliable driving signal for talking-face generation, stronger defense performance should lead to generated results that deviate more from those produced using clean reference inputs. Therefore, larger \textbf{M-LMD} and smaller \textbf{M-SSIM}/\textbf{M-PSNR} indicate stronger disruption of speech-driven facial synthesis, while lower \textbf{CDPAM} indicates higher perceptual similarity between the defended and clean audio.

Compared with \textit{antifake}, our method achieves a substantially lower \textbf{CDPAM} score, indicating that the introduced perturbation is significantly less perceptible in the audio domain. At the same time, it maintains a comparable level of defense effectiveness in terms of \textbf{M-LMD}, \textbf{M-SSIM}, and \textbf{M-PSNR}, showing that our method can effectively interfere with downstream talking-face generation without noticeably compromising perceptual quality.

Although \textit{silencer} achieves the strongest disruption according to \textbf{M-LMD}, \textbf{M-SSIM}, and \textbf{M-PSNR}, this result mainly stems from directly degrading the protected visual input itself. Since the quality of the visual input is severely compromised, the downstream generated videos also deviate more significantly from those driven by clean reference inputs, which naturally leads to stronger disruption metrics. However, this comes at the cost of substantially worse visual appearance and user experience. As also evidenced by the qualitative comparisons in Fig.~\ref{fig:quality}, such a strategy introduces obvious perceptual degradation and produces visually less acceptable protected content.

In contrast, our method achieves effective defense while preserving the natural appearance of the shared visual content. This makes it more suitable for realistic deployment scenarios, where proactive protection should not require sacrificing the viewing experience of benign users. Overall, these results demonstrate that our method provides a more practical trade-off between defense effectiveness and perceptual quality.

\subsection{Qualitative Results}

Figure~\ref{fig:quality} presents qualitative comparisons between our method and representative visual-domain defense approaches \cite{liang2023mist, salman2023PhotoGuard,Salman_2023_advdm, gan2025silence,xue2023toward_effective_protection}. As shown in the figure, existing visual perturbation methods inevitably introduce visible artifacts on the face, including grainy noise, color inconsistency, and corrupted local textures. These distortions are particularly evident in identity-sensitive regions such as the cheeks, lips, and facial contours, which significantly degrade visual realism and user experience.

In contrast, our method achieves \textbf{visually lossless protection}, where the facial appearance remains clean, natural, and free of noticeable artifacts across different identities. The overall texture, color distribution, and facial structure are well preserved, making the defended content perceptually indistinguishable from clean data for human observers.

Despite this high visual fidelity, our method can still effectively disrupt the speech-driven synthesis process, demonstrating that strong protection does not necessarily require sacrificing visual quality. This property is crucial for real-world deployment, where user-uploaded content should remain visually acceptable while being resistant to unauthorized talking-face generation.

\bibliographystyle{ACM-Reference-Format}
\bibliography{sample-base}
\appendix
\clearpage
\section{Appendix}
\label{Appendix}

\subsection{Ablation Experiments}
\begin{table}[t]
\centering
\caption{Ablation study on psychoacoustic masking, band constraint, and smoothing.}
\label{tab:ablation_psy}
\begin{tabular}{c c c c}
\toprule
\textbf{Psycho.} & \textbf{Band} & \textbf{Smooth} & \textbf{CDPAM}$\downarrow$ \\
\midrule
--          & \checkmark & \checkmark & 0.0817 \\
--          & --          & \checkmark & 0.0941 \\
--          & --          & --          & 0.1379 \\
\checkmark  & \checkmark & \checkmark & 0.0634 \\
\bottomrule
\end{tabular}
\end{table}
As shown in Table~\ref{tab:ablation_psy}, each component of our method consistently improves the perceptual quality of the adversarial audio. 
Without any constraint, the perturbation causes the largest perceptual distortion, yielding the worst CDPAM score of 0.1379. 
Introducing temporal smoothing reduces CDPAM to 0.0941, showing that smoothing helps suppress abrupt temporal artifacts. 
Further adding the band constraint improves the score to 0.0817, indicating that constraining perturbations to appropriate frequency regions enhances imperceptibility. 
Finally, incorporating psychoacoustic masking achieves the best result of 0.0634, demonstrating its critical role in reducing perceptually noticeable noise. 
These results verify that the three components are complementary and jointly contribute to more imperceptible adversarial audio.
\begin{algorithm}[t]
\caption{3DMM-Guided Imperceptible Audio Attack for Talking Face Rendering}
\label{alg:3dmm_attack}
\KwIn{Clean waveform $x$}
\KwOut{Adversarial waveform $x^{adv}$}

Compute clean spectrum $S_x=\mathrm{STFT}(x)$ and phase template $P=\exp(j\angle S_x)$\;
Compute psychoacoustic threshold map $\Theta_x$ from $S_x$\;
Compute perturbation bound $C$ from $\Theta_x$\;
Initialize perturbation variable $B$ randomly\;
Initialize threshold Flag $\mathrm{Flag}\leftarrow \mathrm{False}$\;

\For{$t=1$ \KwTo $T$}{
    $A \leftarrow \tanh(B)\odot C$, \quad $\Delta \leftarrow A\odot P$\;
    $x^{adv} \leftarrow \mathrm{ISTFT}(S_x+\Delta)$\;
    $F^{adv}\leftarrow \mathcal{E}(x^{adv})$\;

    $L_{atk}\leftarrow 0$\;
    \ForEach{frame $d_i$}{
        Build audio condition $a_i$ from $F^{adv}$\;
        Render frame $\hat{I}_i \leftarrow \mathcal{R}(d_i,a_i)$\;
        Detect facial landmarks $L_i \leftarrow \mathcal{D}(\hat{I}_i)$\;
        Compute mouth-closure loss $L_{3dmm}^{(i)}$\;
        $L_{atk} \leftarrow L_{atk} + L_{3dmm}^{(i)}$\;
    }
    $L_{atk} \leftarrow \frac{1}{N}L_{atk}$\;

    Compute psychoacoustic masking loss $L_{mask}$\;
    Compute spectral energy regularization $L_{eng}$\;
    Compute temporal-frequency smoothness loss $L_{smooth}$\;

    \tcp{Primary optimization}
    \[
    L_{\mathrm{main}} =
    L_{atk}
    + \lambda_{eng}L_{eng}
    + \lambda_{smooth}L_{smooth}
    + \lambda_{mask}L_{mask}
    \]
    Update $B$ by minimizing $L_{\mathrm{main}}$\;

    \If{$L_{atk}\le\tau$}{
        $\mathrm{flag}\leftarrow \mathrm{True}$\;
        save current $x^{adv}$\;
    }

    \tcp{Secondary refinement after reaching the attack threshold}
    \If{$\mathrm{flag}=\mathrm{True}$ \textbf{and} $L_{atk}\le\tau+\delta$}{
        \[
        L_{\mathrm{psy}}=
        \lambda_{mask}L_{mask}
        + \lambda_{smooth}L_{smooth}
        + \lambda_{reg}\|B\|_{1}
        \]
        Update $B$ by minimizing $L_{\mathrm{psy}}$\;
    }
}
\Return current $x^{adv}$\;
\end{algorithm}

\end{document}